\documentclass{bmvc2k}

\usepackage[misc,geometry]{ifsym}

\newcommand\extrafootertext[1]{%
    \bgroup
    \renewcommand\thefootnote{\fnsymbol{footnote}}%
    \renewcommand\thempfootnote{\fnsymbol{mpfootnote}}%
    \footnotetext[0]{#1}%
    \egroup
}

\usepackage{caption}
\usepackage{cleveref}

\usepackage{listings}
\usepackage{xcolor}

\definecolor{codegreen}{rgb}{0,0.6,0}
\definecolor{codegray}{rgb}{0.95,0.95,0.95}
\definecolor{codepurple}{rgb}{0.58,0,0.82}
\definecolor{backcolour}{rgb}{0.95,0.95,0.92}

\lstdefinestyle{mystyle}{
  backgroundcolor=\color{codegray}, commentstyle=\color{codegreen},
  keywordstyle=\color{magenta},
  numberstyle=\tiny\color{codegray},
  stringstyle=\color{codepurple},
  basicstyle=\ttfamily\bfseries\footnotesize,
  breakatwhitespace=false,         
  breaklines=true,                 
  captionpos=b,                    
  keepspaces=true,                 
  numbers=left,                    
  numbersep=5pt,                  
  showspaces=false,                
  showstringspaces=false,
  showtabs=false,                  
  tabsize=2
}

\newcommand\jsonkey{\color{purple}}
\newcommand\jsonvalue{\color{blue}}
\newcommand\jsonnumber{\color{orange}}

\makeatletter
\newif\ifisvalue@json

\lstdefinelanguage{json}{
    tabsize             = 4,
    showstringspaces    = false,
    keywords            = {false,true},
    alsoletter          = 0123456789.,
    morestring          = [s]{"}{"},
    stringstyle         = \jsonkey\ifisvalue@json\jsonvalue\fi,
    MoreSelectCharTable = \lst@DefSaveDef{`:}\colon@json{\enterMode@json},
    MoreSelectCharTable = \lst@DefSaveDef{`,}\comma@json{\exitMode@json{\comma@json}},
    MoreSelectCharTable = \lst@DefSaveDef{`\{}\bracket@json{\exitMode@json{\bracket@json}},
    basicstyle          = \ttfamily
}

\newcommand\enterMode@json{%
    \colon@json%
    \ifnum\lst@mode=\lst@Pmode%
        \global\isvalue@jsontrue%
    \fi
}

\newcommand\exitMode@json[1]{#1\global\isvalue@jsonfalse}

\lst@AddToHook{Output}{%
    \ifisvalue@json%
        \ifnum\lst@mode=\lst@Pmode%
            \def\lst@thestyle{\jsonnumber}%
        \fi
    \fi
    \lsthk@DetectKeywords%
}

\makeatother

\newcommand{\mycustomsize}{\fontsize{4}{0}\selectfont}

\usepackage{amsmath,amssymb,amsfonts}
\usepackage{mathtools}
\usepackage{commath}
\usepackage{algorithmic}
\usepackage{graphicx}
\usepackage{textcomp}
\usepackage{booktabs}
\definecolor{mygreen}{rgb}{0, 0.56, 0}
\hypersetup{
    hidelinks,
    colorlinks=true,
    citecolor=black,
    urlcolor=Magenta
}

\usepackage{float}

\usepackage{blindtext}

\title{Motion Avatar: Generate Human and Animal Avatars with Arbitrary Motion}

\addauthor{Zeyu Zhang}{https://steve-zeyu-zhang.github.io}{12$*$\dag}
\addauthor{Yiran Wang}{u7079256@anu.edu.au}{2$*$}
\addauthor{Biao Wu}{biaowu165534@gmail.com}{3$*$}
\addauthor{Shuo Chen}{sche0236@student.monash.edu}{4}
\addauthor{Zhiyuan Zhang}{sinns5995@gmail.com}{5}
\addauthor{Shiya Huang}{shiya.huang@student.adelaide.edu.au}{5}
\addauthor{Wenbo Zhang}{wenbo.zhang01@adelaide.edu.au}{5}
\addauthor{Meng Fang}{meng.fang@liverpool.ac.uk}{6}
\addauthor{Ling Chen}{ling.chen@uts.edu.au}{3}
\addauthor{Yang Zhao}{https://yangyangkiki.github.io}{1\text{\Letter}}

\addinstitution{
 La Trobe University\\
 Bundoora VIC, Australia
}
\addinstitution{
 The Australian National University\\
 Canberra ACT, Australia
}
\addinstitution{
 University of Technology Sydney\\
 Ultimo NSW, Australia
}
\addinstitution{
 Monash University\\
 Clayton VIC, Australia
}
\addinstitution{
 The University of Adelaide\\
 Adelaide SA, Australia
}
\addinstitution{
 The University of Liverpool\\
 Liverpool, United Kingdom\\
 \vspace{\baselineskip}
 \fontsize{8}{15}\selectfont $^{*}$Equal contribution.\\ 
 \fontsize{8}{15}\selectfont $^{\text{\Letter}}$Corresponding author.\\
 \fontsize{8}{15}\selectfont $^\dag$Work done while being a research assistant\\ 
 at La Trobe University.}

\runninghead{Zhang et al.}{Motion Avatar}

\begin{document}

\maketitle

\begin{center}
\vspace{-0.6cm}
 \centering
 Project Website: \href{https://steve-zeyu-zhang.github.io/MotionAvatar}{https://steve-zeyu-zhang.github.io/MotionAvatar}
\end{center}

\begin{abstract}

In recent years, there has been significant interest in creating 3D avatars and motions, driven by their diverse applications in areas like film-making, video games, AR/VR, and human-robot interaction. However, current efforts primarily concentrate on either generating the 3D avatar mesh alone or producing motion sequences, with integrating these two aspects proving to be a persistent challenge. Additionally, while avatar and motion generation predominantly target humans, extending these techniques to animals remains a significant challenge due to inadequate training data and methods. To bridge these gaps, our paper presents three key contributions. Firstly, we proposed a novel agent-based approach named \textbf{Motion Avatar}, which allows for the automatic generation of high-quality customizable human and animal avatars with motions through text queries. The method significantly advanced the progress in dynamic 3D character generation. Secondly, we introduced a \textbf{LLM planner} that coordinates both motion and avatar generation, which transforms a discriminative planning into a customizable Q\&A fashion. Lastly, we presented an animal motion dataset named \textbf{Zoo-300K}, comprising approximately \textbf{300,000} text-motion pairs across \textbf{65} animal categories and its building pipeline \textbf{ZooGen}, which serves as a valuable resource for the community. 

\end{abstract}

\vspace{-0.7cm}
\section{Introduction}

The field of computational modeling for dynamic 3D or 4D avatars holds significant importance across various domains such as robotics, virtual, and augmented reality, computer gaming, and multimedia. Creating high-quality and user-friendly animated avatars is a desired goal of the 3D computer vision community. This involves not only ensuring the visual appeal of the avatars but also prioritizing their functionality and ease of use. Traditional methods \cite{yin20234dgen, gao2024gaussianflow, liu2024dynamic} involve extracting information directly from video recordings, then using the data to model and reconstruct dynamic avatars in both spatial and temporal dimensions. Other approaches \cite{zhao2023animate124} involve integrating 3D reconstruction with video diffusion to bring 3D meshes to life through animation. However, these methods often suffer from imprecise and limited motion control, or exhibit inconsistencies between multiple views of the 3D mesh. These limitations hinder the seamless manipulation and other real world applications of dynamic avatars in interactive settings.

Recent advancements in \textit{text-to-motion} generation \cite{guo2022generating,tevet2022human,chen2023executing,zhang2024motion,guo2023momask}, employing both diffusion \cite{ho2020denoising, song2020denoising, dhariwal2021diffusion, nichol2021glide, rombach2022high} and autoregressive \cite{vaswani2017attention, radford2018improving} models, have demonstrated significant promise. These advancements represent a notable step forward in generating motion sequences directly from textual descriptions, which brings a new approach to dynamic motion avatar generation. Meanwhile, significant advancements in 3D mesh reconstruction and generation \cite{qian2023magic123,liu2023zero,liu2024one,hong2023lrm,tochilkin2024triposr}, particularly through approach of \textit{image-to-3D}, have demonstrated considerable promise and dominance in the field of 3D object reconstruction. These advancements hold great potential for real-world applications. However, existing efforts predominantly focus on either creating the 3D avatar mesh independently or generating motion sequences separately. Yet, effectively integrating these two components remains a persistent challenge.

Moreover, although avatar and motion generation techniques are primarily developed for human characters, their adaptation to animals presents a formidable obstacle. This challenge arises primarily from the scarcity of suitable training data and the limitations of existing methodologies. Expanding these technologies to encompass animal characters requires novel approaches and a deeper understanding of animal behavior and anatomy \cite{yang2023omnimotiongpt}. 

Therefore, in order to address these challenges, our study delineates three principal contributions as follows:

\begin{itemize}
    \item Firstly, we introduced a groundbreaking method named \textbf{Motion Avatar}, which leverages LLM agent-based techniques to automatically create customizable human and animal avatars complete with dynamic motions based on text input. This innovation represents a significant leap forward in the field of dynamic 3D character generation, offering high-quality avatars tailored to specific needs and preferences. With Motion Avatar, users can effortlessly generate lifelike characters with realistic movements, pushing the boundaries of what's achievable in avatar creation through textual condition.
    \item Secondly, we introduced a \textbf{LLM planner} that handles both motion and avatar generation. This planner takes a discriminative planning approach and refines it into a more flexible Q\&A paradigm. The generalization ability of the LLM planner will help adapt to broader dynamic avatar generation tasks in the future.
    \item Lastly, we presented a comprehensive animal motion dataset called \textbf{Zoo-300K}, which contains around \textbf{300,000} pairs of text descriptions and corresponding motion data covering \textbf{65} different animal categories. Additionally, we developed \textbf{ZooGen}, a robust pipeline for constructing such datasets, which stands as a valuable asset for the wider research community.
\end{itemize}

\vspace{-0.7cm}
\section{Related Works}

\paragraph{Text-to-Motion Generation}

Creating motion is a vital aspect of computer vision, playing a crucial role in diverse applications such as 3D video animation, computer games, and robot manipulation. A prevalent approach to generating motion, known as the \textit{Text-to-Motion} task, revolves around establishing a shared latent space for both textual descriptions and motion data. Among these applications, human motion generation has garnered significant attention and exploration.
MotionCLIP \cite{tevet2022motionclip} leverages Transformer-based \cite{vaswani2017attention} Autoencoders to reconstruct motion sequences while ensuring alignment with corresponding textual labels within the CLIP \cite{radford2021learning} space. This alignment effectively integrates semantic knowledge from CLIP into the manifold of human motion. TEMOS \cite{petrovich2022temos} and T2M \cite{guo2022generating} merge a Transformer-based Variational Autoencoder (VAE) \cite{kingma2013auto} with a text encoder to produce distribution parameters that operate within the latent space of the VAE. AttT2M \cite{zhong2023attt2m} and TM2D \cite{gong2023tm2d} integrate a body-part spatio-temporal encoder into VQ-VAE to facilitate improved learning of a discrete latent space with heightened expressiveness. 
MotionDiffuse \cite{zhang2024motiondiffuse} presents a pioneering framework for generating motion guided by text, leveraging diffusion models \cite{ho2020denoising, song2020denoising, dhariwal2021diffusion, nichol2021glide, rombach2022high}. It demonstrates various appealing traits such as probabilistic mapping, lifelike synthesis, and multi-tiered manipulation. Conversely, MDM \cite{tevet2022human} employs a Transformer-based diffusion model devoid of classifiers to forecast samples, rather than noise, at each step. In contrast, MLD \cite{chen2023executing} conducts diffusion within latent motion space, eschewing the utilization of a diffusion model to forge links between raw motion sequences and conditional inputs. Motion Mamba \cite{zhang2024motion} and InfiniMotion \cite{zhang2024infinimotion} proposed approaches which leveraged state space models (SSMs) \cite{gu2023mamba} and diffusion to design an efficient architecture for long-sequence human motion generation. 
MoMask \cite{guo2023momask} proposed a novel masked modeling framework, designed a hierarchical quantization scheme architecture, and achieved superior results on text-driven 3D human motion generation.

Currently, generating animal motion possesses greater challenges compared to human motion generation. These challenges primarily stem from inadequate data and inconsistent motion representation, such as variations in the number of joints and kinematics. Nevertheless, there are ongoing efforts aimed at addressing these hurdles. OmniMotionGPT \cite{yang2023omnimotiongpt} proposes a method for generating diverse and realistic animal motion sequences from textual descriptions, achieving superior results compared to traditional human motion synthesis approaches, by leveraging a model architecture inspired by GPT \cite{radford2018improving,radford2019language,brown2020language,achiam2023gpt} and joint training of motion autoencoders for both animal and human motions. 
SinMDM \cite{raab2023single} introduced a method for synthesizing diverse motions of arbitrary length from a single input motion sequence, addressing the scarcity of data instances in animating animals and exotic creatures, and outperformed existing methods in quality and efficiency.

\paragraph{Avatar Generation}

Previously, research on 3D avatar and texture generation has drawn inspiration from text-guided 2D image generation techniques. Typically, these methods involve either training a normalizing flow model based on textual descriptions or employing a coarse-to-fine pipeline to create detailed 3D textured meshes. CLIP-Mesh \cite{mohammad2022clip} introduced a novel method for zero-shot 3D model generation solely from textual prompts, achieving impressive results without the need for 3D supervision. AvatarCLIP \cite{hong2022avatarclip} introduced a text-driven framework for 3D avatar generation and animation, utilizing CLIP \cite{radford2021learning} to enable layman users to customize avatars and animate them using natural language descriptions. Fantasia3D \cite{chen2023fantasia3d} introduces a novel method for high-quality text-to-3D content creation, emphasizing disentangled modeling of geometry and appearance, which outperforms existing methods in various task settings. DreamFusion \cite{poole2022dreamfusion} introduces a novel method leveraging a pretrained 2D text-to-image diffusion model for text-to-3D synthesis, achieving impressive results without the need for large-scale 3D datasets or modifications to the image diffusion model. ZeroAvatar \cite{weng2023zeroavatar} introduces a method for improving optimization-based image-to-3D avatar generation by incorporating an explicit 3D human body prior, resulting in enhanced robustness and 3D consistency compared to existing zero-shot methods. NIA \cite{kwon2022neural} proposes a method for synthesizing varied views and poses of humans from sparse multi-view images, outperforming existing methods in both identity and pose generalization. DreamAvatar \cite{cao2023dreamavatar} proposed a text-and-shape guided framework, achieving superior performance in generating high-quality 3D human avatars with controllable poses. XAGen \cite{xu2024xagen} proposed a novel method for 3D human avatar generation with expressive control over body, face, and hands, surpassing existing state-of-the-art methods in realism, diversity, and control abilities.

Recently, there's been a promising trend in 3D generation solely from a single image, exemplified by works like Magic123 \cite{qian2023magic123}, Zero123 \cite{liu2023zero, shi2023zero123++}, One-2-3-45 \cite{liu2024one, liu2023one}, and LRM \cite{hong2023lrm}. The most commonly used approach for single-image-to-3D conversion is based on the LRM paradigm, which employs an image-to-triplane decoder along with a triplane NeRF \cite{chan2022efficient} or Gaussian Splatting \cite{kerbl20233d} technique. Notable examples include TripoSR \cite{tochilkin2024triposr}, CRM \cite{wang2024crm}, TriplaneGaussian \cite{zou2023triplane}, and LGM \cite{tang2024lgm}.

\paragraph{LLM-powered Agent} LLM agents utilize large language models as a central component to simulate human-like decision-making processes, playing a crucial role in the progression towards artificial general intelligence. Generative Agents \cite{park2023generative} introduced a method to simulate believable human behavior, showcasing it in an interactive sandbox environment where agents autonomously plan and engage in social interactions. Voyager \cite{wang2023voyager} introduced as the first LLM-powered embodied lifelong learning agent in Minecraft, achieving exceptional proficiency in exploration, skill acquisition, and novel discovery without human intervention. LLM-Planner \cite{song2023llm} introduced a novel approach to few-shot planning for embodied agents, achieving competitive performance with minimal training data, thus enabling the development of versatile and sample-efficient agents capable of completing complex tasks. ELLM \cite{du2023guiding} introduced a method that leverages language model pretraining to reward agents for achieving goals suggested by a language model prompted with a description of the agent's current state, which typically improves performance on downstream tasks. GLAM \cite{carta2023grounding} introduced functional grounding to align large language models with the environment, achieving improved decision-making capabilities.

\vspace{-0.7cm}
\section{Datasets}

\paragraph{Zoo-300K and ZooGen}

\begin{figure}
    \centering
    \includegraphics[width=\linewidth]{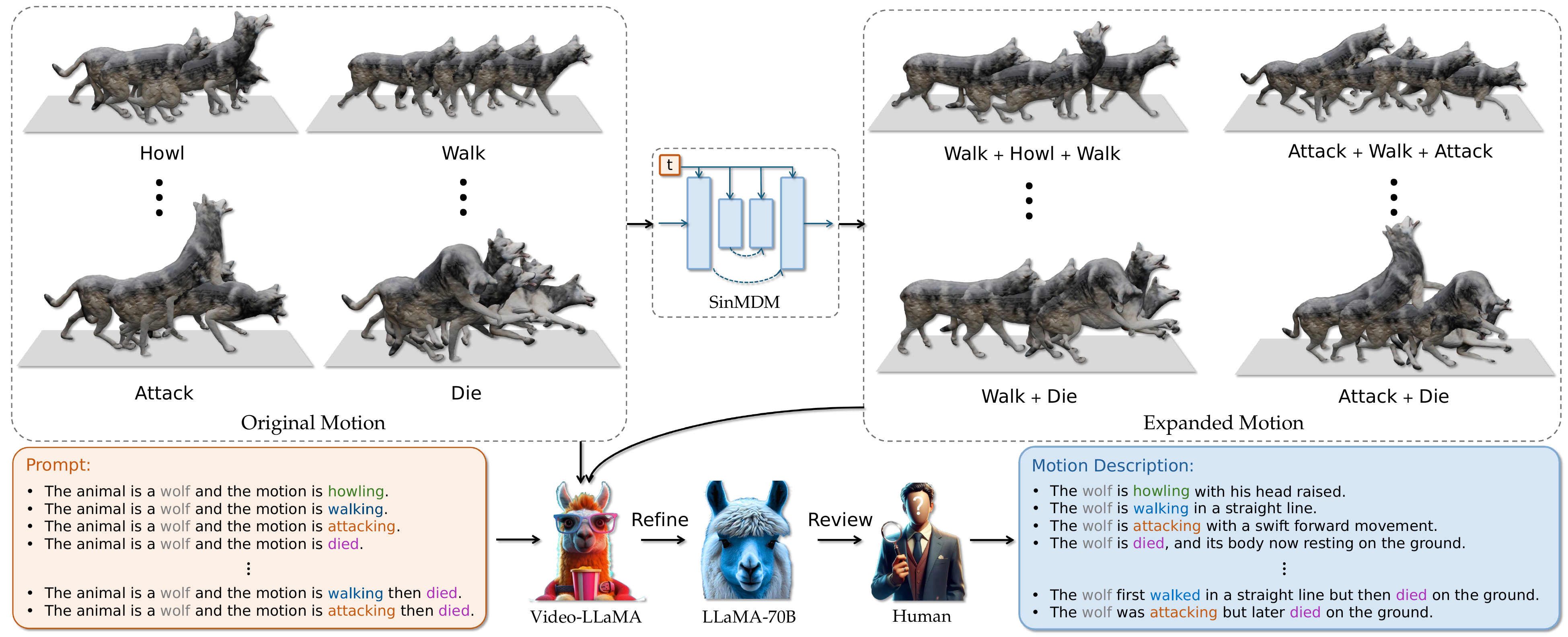} 
    \vspace{0.5cm}
    \caption{The diagram illustrates the process of our proposed \textbf{ZooGen}. Initially, SinMDM \cite{raab2023single} is employed to edit and enhance motion within Truebones Zoo \cite{free}. Subsequently, Video-LLaMA \cite{zhang2023video} is utilized to describe the motion in a paragraph, followed by refinement using LLaMA-70B \cite{touvron2023llama}. Finally, human review is conducted on the motion captions, which are then gathered as textual descriptions in the \textbf{Zoo-300K} dataset.}
    \label{fig:zoogen}
\end{figure}

Presently, a major obstacle in generating animal motion is the lack of sufficient data pairs containing both animal motion and text descriptions. Compared to generating human motion, generating animal motion based on text commands hasn't been explored as extensively, mostly because there's a scarcity of suitable datasets. Animal motion datasets in research are severely limited and nowhere near as extensive as those available for human motion. To address this gap, we introduced \textbf{Zoo-300K}, a dataset containing around \textbf{300,000} pairs of text descriptions and corresponding animal motions spanning \textbf{65} different animal categories. Based on the Truebones Zoo \cite{free} dataset, which comprises human-cultivated animal motion data annotated with textual labels denoting animal and motion categories, we have introduced a dataset construction pipeline named \textbf{ZooGen}, which facilitates the creation of a text-driven animal motion dataset, shown in \cref{fig:zoogen}.

ZooGen utilizes SinMDM \cite{raab2023single} to enhance motion editing and expansion capabilities, which is training a motion diffusion model on a single motion. This model can then be applied for editing and combining with other motions effectively. First, we utilize the animal motions in Truebones Zoo \cite{free}, which are cultivated by humans. For each motion, we train a unique diffusion model. These models are then used to modify other animal motions in various permutations and combinations, enabling us to acquire all the animal motions in our dataset, as shown in the upper section of \cref{fig:zoogen}.

While human motion captioning has made strides \cite{radouane2024motion2language,guo2022tm2t}, accurately captioning animal motion remains a significant challenge. To address this gap, we employed a multimodal large language model Video-LLaMA \cite{zhang2023video}, specifically designed for comprehensive video understanding, to describe our refined motion data. To improve the clarity and quality of captions, we use LLaMA-70B \cite{touvron2023llama} for additional refinement. Eventually, these captions will undergo manual review and be selected for inclusion in the final text of Zoo-300K, as shown in the latter section of \cref{fig:zoogen}.

\paragraph{HumanML3D} For human motion generation, we used the go-to dataset HumanML3D \cite{guo2022generating}, which combines 14,616 motions collected from the AMASS \cite{mahmood2019amass} and HumanAct12 \cite{guo2020action2motion} datasets. Each motion in the dataset is described by three text descriptions, totaling 44,970 scripts. It covers various actions like exercising, dancing, and acrobatics, providing a diverse collection of motion with associated language descriptions.

\paragraph{Avatar Q\&A Dataset}

For tuning and evaluating our LLM planner, we've created the avatar Q\&A dataset. This dataset utilizes animal and motion categories from Zoo-300K as input and is generated by LLaMA-70B \cite{touvron2023llama}. It serves as a corpus for adjusting our LLM planner, with the \textit{instruction} attribute used as prompts and the \textit{output} attribute as the desired output. The dataset contains 1756 items, and Listing \ref{lst:qa} in Appendix \ref{app:c} provides several examples. The dataset will assess whether the LLM planner can accurately identify animal and motion categories from natural language descriptions provided by users.

\vspace{-0.5cm}
\section{Methodology}

\paragraph{LLM Planner}

\begin{figure}
    \centering
     \includegraphics[width=1\textwidth]{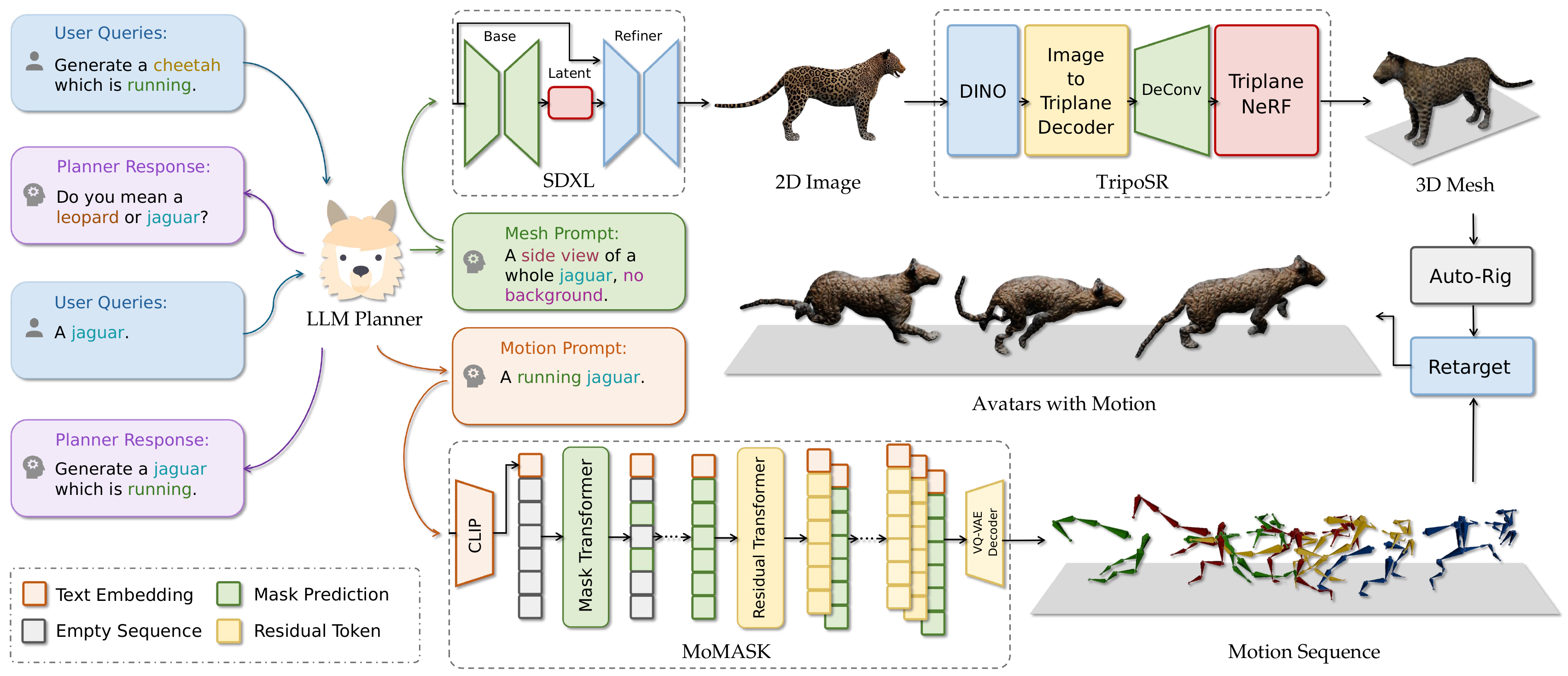}
     \vspace{0.1cm}
     \caption{\textbf{Motion Avatar} utilizes a LLM-agent based approach to manage user queries and produce tailored prompts. These prompts are designed to facilitate both the generation of motion sequences and the creation of 3D meshes. Motion generation follows an autoregressive process, while mesh generation operates within an image-to-3D framework. Subsequently, the generated mesh undergoes an automatic rigging process, allowing the motion to be retargeted to the rigged mesh.}
      \label{fig:main}
\end{figure}

The objective of LLM planner design is to efficiently and automatically extract useful information from the user prompt $Q_{U}$, then use external components to create customizable animated avatars effectively. To efficiently deploy the planner on a local device, we developed the LLM planner leveraging the LLaMA-7B framework \cite{touvron2023llama}. This planner was finely tuned using the avatar Q\&A dataset to better suit the task of generating motion avatars.

In contrast to traditional text classification tasks that focus on classifying semantics within specific domains, we've developed our planner through instruction tuning. This approach effectively adapts the planner to our needs and activates the potential of large language model to identify motion $M$ and avatar $A$ categories within the dialogue between the user and the planner. Furthermore, the LLM planner possesses the ability to generate corresponding prompt $Q_{M}$ and $Q_{A}$ for downstream avatar motion and mesh generation, which can be formulated as:

\vspace{-0.3cm}
\begin{equation}
    (Q_{M}, Q_{A}) = \text{LLM-Planner}(Q_{U})
\end{equation}

Eventually, the LLM planner will be able to recognize main subjects in the dialogue beyond avatars and motion categories within Zoo-300K. The domain generalization capability of our LLM planner will help adapt to more generalizable animal motion generation tasks for the future. Meanwhile, the LLM planner effectively coordinates the avatar and motion generation components to generate personalized results that meet the users' needs, as shown in \cref{fig:main}.

\begin{figure}
    \centering
    \includegraphics[width=\linewidth]{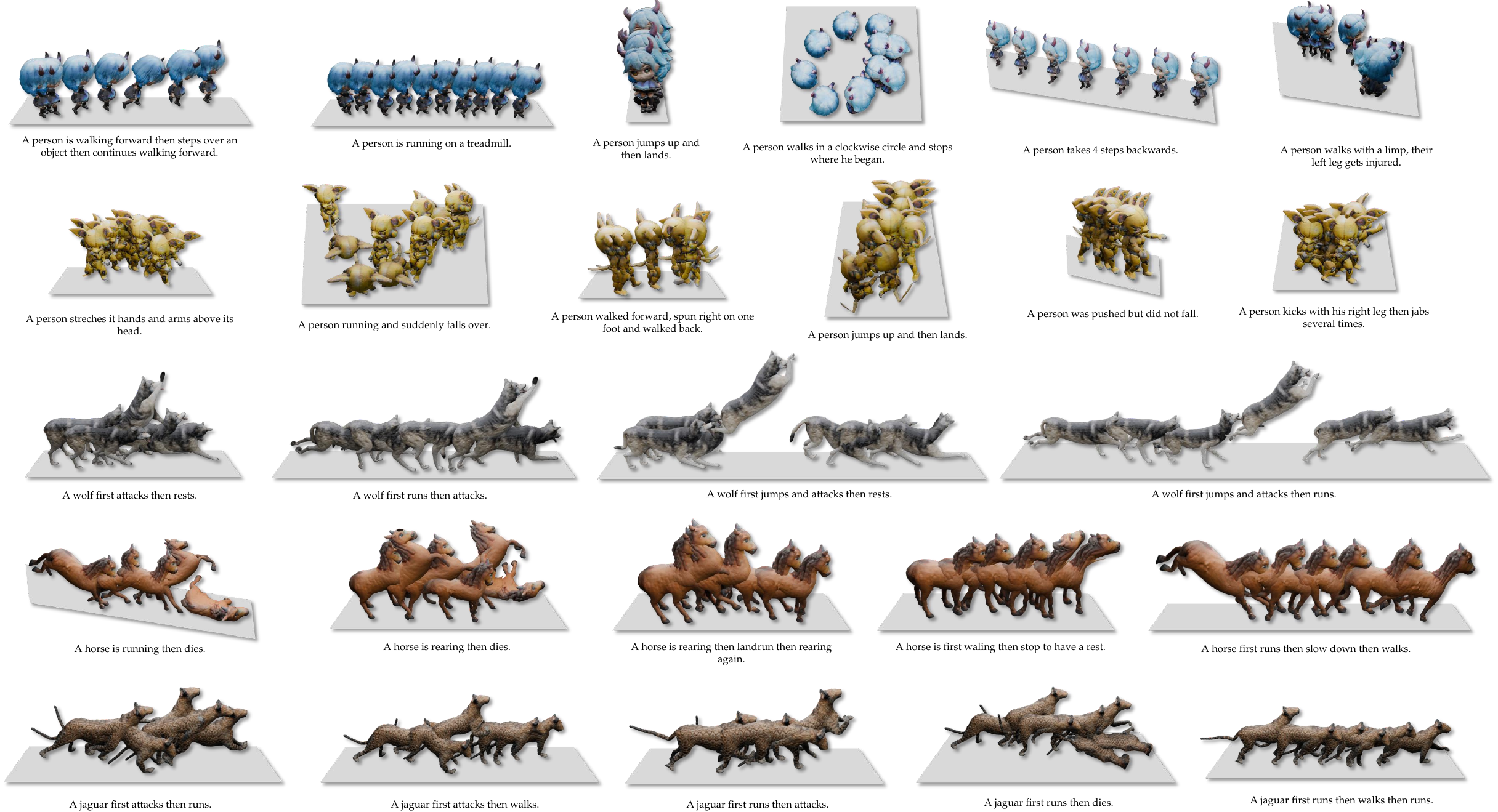} 
    \vspace{0.5cm}
    \caption{The figure illustrates various examples of animal motion generated by Motion Avatar, demonstrating its ability to produce high-quality motion and mesh for both human and animal characters.}
    \label{fig:motion}
\end{figure}

\paragraph{Motion Generation}

Motion generation with MoMask \cite{guo2023momask} involves a two-stage training process. Firstly, a residual VQ-VAE compresses a motion sequence $\mathbf{m}_{1:N} \in \mathbb{R}^{N \times D}$ into a latent vector $\mathbf{\tilde{b}}_{1:n} \in \mathbb{R}^{n \times d}$, which is then replaced with its nearest code entry in a codebook $C = \{{\mathbf{c}_k}\}^K_{k=1} \subset \mathbb{R}^d$. Then the decoder projects the quantized code sequence $\mathbf{b}_{1:n} = Q(\mathbf{\tilde{b}}_{1:n}) \in \mathbb{R}^{n \times d}$ back to motion space for motion reconstruction $\mathbf{\hat{m}} = D(\mathbf{b})$. 

In the second stage, both the mask transformer and the residual transformer are trained simultaneously. The text embedding $c$ is extracted from prompt $Q_M$ using CLIP \cite{radford2021learning}, and the mask transformer draws inspiration from the bidirectional mask mechanism found in BERT \cite{devlin2018bert}. Initially, the mask transformer randomly replaced a varying portion of sequence elements $t^0_{1:n} \in \mathbb{R}^n$ with a special [MASK] token, aiming to predict these masked tokens based on text embedding $c$ and sequence after masking $\tilde{t}^0$. Meanwhile, the residual transformer is trained to model tokens from other $V$ residual quantization layers. During training, a quantizer layer $j \in [1,V]$ is randomly selected, and all tokens in the preceding layers $t^{0:j-1}$ are embedded and summed up as the token embedding input. Using token embedding, text embedding, and the RQ layer indicator $j$ as input, the residual transformer $p_{\phi}$ is trained to predict the $j-th$ layer tokens simultaneously.

During inference, the mask transformer is given an empty motion sequence $t^0(0)$ as an input, and expected to generate the base-layer token sequence $t^0$ of length $n$ in $L$ iterations. Then the residual transformer progressively predicts the token sequence in the rest quantization layers. Eventually, the latent motion embedding will be decoded and projected back to motion space through the decoder of residual VQ-VAE, as shown in \cref{fig:main}.

\paragraph{Avatar Generation}

For generation of avatar meshes, the Motion Avatar initially employed Stable Diffusion XL \cite{podell2023sdxl} to utilize the mesh prompt $Q_A$ generated by LLM planner, producing a 2D image $I_A$ of the avatar's front or side view tailored to requirements. The process could be described as

\vspace{-0.3cm}
\begin{equation}
    I_A = \text{SDXL}(Q_A)
\end{equation}

Given a 2D input $I_A$, TripoSR \cite{tochilkin2024triposr} initially obtains the image representation using a DINO \cite{caron2021emerging, oquab2023dinov2} encoder, followed by the image-to-triplane decoder that transforms the image embedding into the triplane-NeRF \cite{chan2022efficient} representation. This decoder comprises transformer layers, each featuring a self-attention layer for attending to different parts of the triplane representation and learning their relationships, and a cross-attention layer for incorporating global and local image features into the triplane representation. The triplane-NeRF \cite{chan2022efficient} representation serves as a compact and expressive 3D representation, ideal for depicting objects with intricate shapes and textures. Finally, the NeRF \cite{mildenhall2021nerf} model consists of a stack of MLPs \cite{rumelhart1986learning}, responsible for predicting the color and density of a 3D point in space and rendering into a 3D mesh. The avatar mesh will be later rigged by Auto-Rig \cite{autorig}, enabling the motion to
be retargeted to the rigged mesh, as shown in \cref{fig:main}.

\vspace{-0.3cm}
\section{Experiments}

\subsection{LLM Planner Evaluation}

We extensively evaluate our LLM planner using instruction tuning on the Avatar Q\&A Dataset, compared with zero-shot LLaMA-7B \cite{touvron2023llama}. The results in the \cref{table:llm} demonstrate that our LLM planner successfully recognizes motion and avatar categories within the dialogue between the user and the planner, showing promising potential for generalization ability in coordinating various dynamic avatar tasks. This thorough evaluation underscores the robustness and versatility of our approach in the context of dynamic avatar generation.

\begin{table}[h]
\centering
\resizebox{0.7\textwidth}{!}{
\begin{tabular}{c||cc|c}
\toprule
Models & Animal Acc. $\uparrow$ & Motion Acc. $\uparrow$ & Overall Acc. $\uparrow$ \\ \midrule 
LLaMA-7B \cite{touvron2023llama} & $69.30$ & $19.19$ & $44.24$ \\
\textbf{LLM Planner (Ours)} & $\textbf{97.07\textcolor{mygreen}{\raisebox{-0.8ex}{\textsuperscript{ +27.77}}}}$ & $\textbf{71.67\textcolor{mygreen}{\raisebox{-0.8ex}{\textsuperscript{ +52.48}}}}$ & $\textbf{84.37\textcolor{mygreen}{\raisebox{-0.8ex}{\textsuperscript{ +40.13}}}}$ \\
\midrule
\end{tabular}
}
\vspace{0.5cm}
\caption{Evaluation of our \textbf{LLM Planner} against zero-shot LLaMA-7B \cite{touvron2023llama} on the Avatar Q\&A Dataset reveals its successful recognition of motion and avatar categories during user-planner dialogue.}
\label{table:llm}
\vspace{-1cm}
\end{table}

\subsection{Motion Evaluation}

Since human motion generation has already been evaluated in the MoMask \cite{guo2023momask}, our focus shifts to comprehensively assessing animal motion generation utilizing the newly introduced Zoo-300K dataset. For quantitative results, please refer to the table in \textbf{Appendix \ref{app:a}: Animal Motion Quantative Evaluation} and \textbf{Appendix \ref{app:b}: User Study} in supplementary. For qualitative assessment, the visualization in \cref{fig:motion} illustrates our model's capability to seamlessly generate animal motion based on text conditions, showcasing potential applications in fields like computer gaming and film making. These findings highlight the versatility and effectiveness of our approach in diverse scenarios.

\vspace{-0.2cm}
\subsection{Avatar Evaluation}

\begin{figure}
    \centering
    \includegraphics[width=\linewidth]{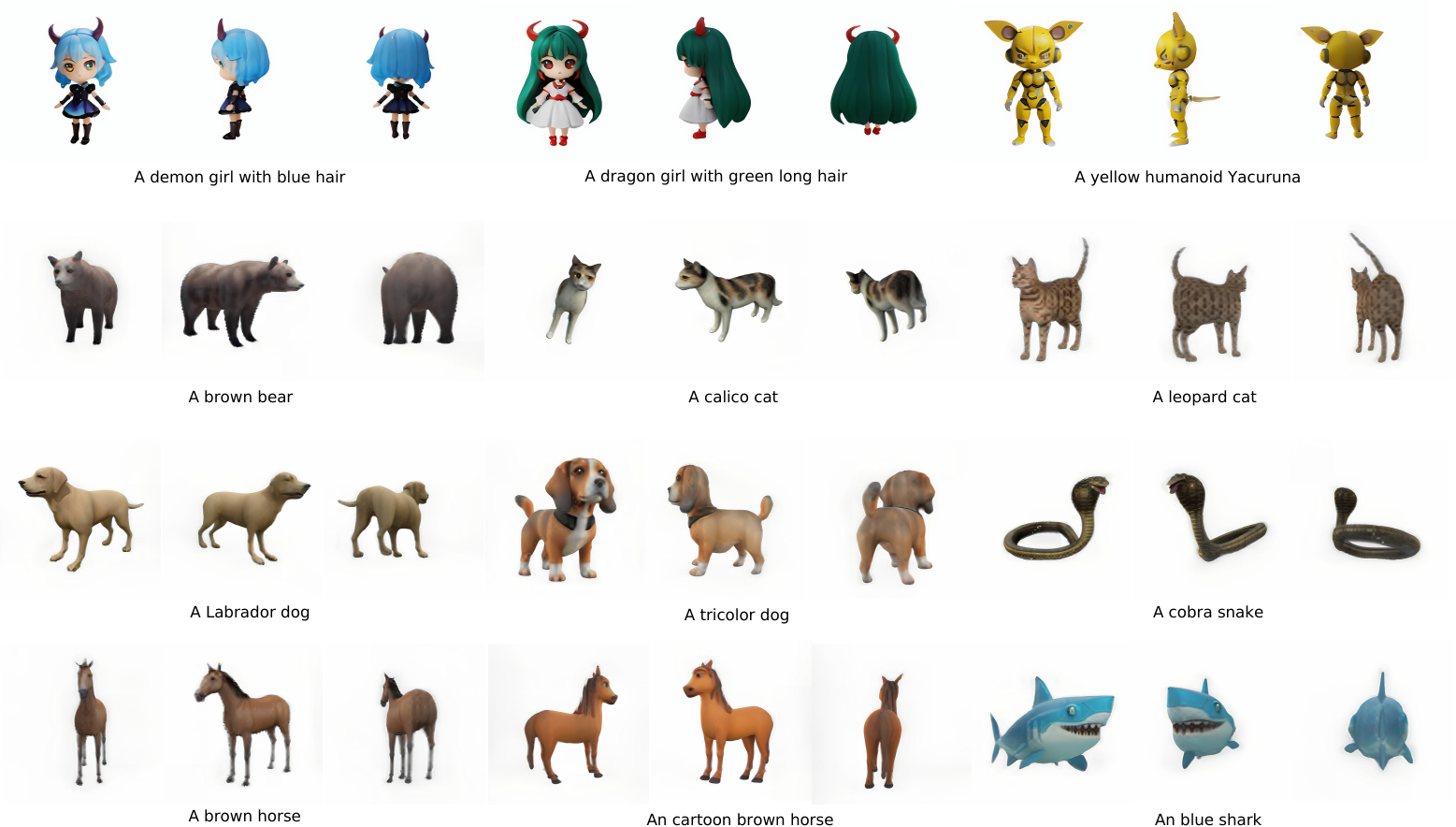} 
    \vspace{0.5cm}
    \caption{The figure showcases various examples of generated 3D meshes, encompassing both human and animal avatars. The meshes exhibit high-quality geometry and offer customizable textures, thus serving as a robust foundation for avatar animation. Furthermore, this advancement holds promise for enhancing the technique's applicability in real-world scenarios.}
    \label{fig:avatar}
\end{figure}

Given that TripoSR \cite{tochilkin2024triposr} has already conducted quantitative evaluations for 3D mesh generation, we extended our evaluation to generating 3D avatars using the character categories in the Zoo-300K dataset to assess our model's adaptation to our task. The \cref{fig:avatar} illustrates randomly selected examples of 3D avatars generated by our model. The results demonstrate that our model can produce high-quality and detailed 3D avatars, indicating its utility for avatar motion generation.

\vspace{-0.4cm}
\section{Discussion and Conclusion}

In conclusion, our study addresses the ongoing challenges in dynamic 3D avatar generation by presenting innovative solutions. Through the introduction of Motion Avatar, a novel agent-based approach, we enable the automatic creation of highly customizable human and animal avatars complete with dynamic motions based on text queries, thus advancing the field significantly. Additionally, our LLM planner facilitates the coordination of motion and avatar generation, enhancing adaptability and usability in dynamic avatar tasks. Furthermore, the development of the Zoo-300K dataset, alongside the ZooGen pipeline, provides a valuable resource for researchers, underscoring our commitment to advancing the field of dynamic avatar generation across various domains.

\clearpage

\appendix
\section*{Appendix}
\section{Animal Motion Quantative Evaluation}
\label{app:a}


We evaluated our Motion Avatar on the Zoo-300K dataset, and the results are presented in \cref{tab:overall}. The comprehensive evaluation demonstrated that our method achieves high-quality animal motion generation. The results indicate that our approach produces highly realistic and promising animations, highlighting the effectiveness and potential of our technique in generating detailed and lifelike animal motions.

\begin{table}[htbp]
\centering
\caption{The table shows that our \textbf{Motion Avatar} achieves high-quality animal motion generation on the Zoo-300K dataset. It demonstrates the effectiveness of our method, producing realistic and promising animal motions.}
\vspace{0.5cm}
\resizebox{0.8\columnwidth}{!}{%
\begin{tabular}{lcccccc}
\toprule
Method & R-Prec Top 1 $\uparrow$ & R-Prec Top 2 $\uparrow$ & R-Prec Top 3 $\uparrow$ & FID $\downarrow$ & MultiModal-Dist $\downarrow$ & Diversity $\rightarrow$\\ 
\midrule
Anaconda (GT) & $0.400$ & $0.591$ & $0.713$ & $0.015$ & $2.870$ & $10.233$ \\
\midrule
\textbf{Anaconda (Ours)} & $\mathbf{0.042}$ & $\mathbf{0.080}$ & $\mathbf{0.128}$ & $\mathbf{0.51067}$ & $\mathbf{9.053}$ & $\mathbf{5.165}$ \\
\midrule
Ant (GT) & $0.413$ & $0.650$ & $0.768$ & $0.12$ & $2.000$ & $9.700$ \\
\midrule
\textbf{Ant (Ours)} & $\mathbf{0.056}$ & $\mathbf{0.110}$ & $\mathbf{0.166}$ & $\mathbf{68.220}$ & $\mathbf{9.380}$ & $\mathbf{4.380}$ \\
\midrule
Bat (GT) & $0.406$ & $0.677$ & $0.841$ & $0.011$ & $1.805$ & $12.084$ \\
\midrule
\textbf{Bat (Ours)} & $\mathbf{0.193}$ & $\mathbf{0.536}$ & $\mathbf{0.086}$ & $\mathbf{79.403}$ & $\mathbf{9.430}$ & $\mathbf{0.984}$ \\
\midrule
Bear (GT) & $0.641$ & $0.851$ & $0.916$ & $0.012$ & $1.827$ & $13.587$ \\
\midrule
\textbf{Bear (Ours)} & $\mathbf{0.156}$ & $\mathbf{0.287}$ & $\mathbf{0.382}$ & $\mathbf{35.735}$ & $\mathbf{8.890}$ & $\mathbf{4.364}$ \\
\midrule
Bird (GT) & $0.569$ & $0.762$ & $0.851$ & $0.004$ & $2.679$ & $12.294$ \\
\midrule
\textbf{Bird (Ours)} & $\mathbf{0.059}$ & $\mathbf{0.091}$ & $\mathbf{0.127}$ & $\mathbf{54.735}$ & $\mathbf{9.726}$ & $\mathbf{4.389}$ \\
\midrule
Buffalo (GT) & $0.547$ & $0.750$ & $0.852$ & $0.014$ & $3.876$ & $16.405$ \\
\midrule
\textbf{Buffalo (Ours)} & $\mathbf{0.070}$ & $\mathbf{0.140}$ & $\mathbf{0.227}$ & $\mathbf{167.900}$ & $\mathbf{13.378}$ & $\mathbf{4.352}$ \\
\midrule
Buzzard (GT) & $0.423$ & $0.639$ & $0.778$ & $0.007$ & $2.654$ & $11.045$ \\
\midrule
\textbf{Buzzard (Ours)} & $\mathbf{0.034}$ & $\mathbf{0.084}$ & $\mathbf{0.111}$ & $\mathbf{65.274}$ & $\mathbf{8.645}$ & $\mathbf{1.824}$ \\
\midrule
Camel (GT) & $0.297$ & $0.469$ & $0.609$ & $0.014$ & $3.381$ & $11.025$ \\
\midrule
\textbf{Camel (Ours)} & $\mathbf{0.078}$ & $\mathbf{0.188}$ & $\mathbf{0.250}$ & $\mathbf{92.963}$ & $\mathbf{11.654}$ & $\mathbf{5.249}$ \\
\midrule
Cat (GT) & $0.141$ & $0.281$ & $0.422$ & $0.041$ & $2.274$ & $9.060$ \\
\midrule
\textbf{Cat (Ours)} & $\mathbf{0.063}$ & $\mathbf{0.188}$ & $\mathbf{0.219}$ & $\mathbf{56.405}$ & $\mathbf{7.195}$ & $\mathbf{1.969}$ \\
\midrule
Centipede (GT) & $0.386$ & $0.580$ & $0.712$ & $0.026$ & $2.256$ & $11.279$ \\
\midrule
\textbf{Centipede (Ours)} & $\mathbf{0.123}$ & $\mathbf{0.223}$ & $\mathbf{0.317}$ & $\mathbf{65.076}$ & $\mathbf{8.590}$ & $\mathbf{5.664}$ \\
\midrule
Chicken (GT) & $0.094$ & $0.203$ & $0.313$ & $0.075$ & $3.503$ & $9.196$ \\
\midrule
\textbf{Chicken (Ours)} & $\mathbf{0.063}$ & $\mathbf{0.141}$ & $\mathbf{0.172}$ & $\mathbf{70.271}$ & $\mathbf{7.953}$ & $\mathbf{2.752}$ \\
\midrule
Cobra (GT) & $0.386$ & $0.583$ & $0.700$ & $0.009$ & $2.881$ & $10.901$ \\
\midrule
\textbf{Cobra (Ours)} & $\mathbf{0.217}$ & $\mathbf{0.324}$ & $\mathbf{0.406}$ & $\mathbf{0.309}$ & $\mathbf{6.397}$ & $\mathbf{10.823}$ \\
\midrule
Komodo (GT) & $0.260$ & $0.375$ & $0.456$ & $0.064$ & $5.208$ & $9.968$ \\
\midrule
\textbf{Komodo (Ours)} & $\mathbf{0.073}$ & $\mathbf{0.125}$ & $\mathbf{0.161}$ & $\mathbf{43.677}$ & $\mathbf{7.074}$ & $\mathbf{3.047}$ \\
\midrule
Coyote (GT) & $0.523$ & $0.765$ & $0.869$ & $0.034$ & $2.042$ & $12.985$ \\
\midrule
\textbf{Coyote (Ours)} & $\mathbf{0.052}$ & $\mathbf{0.102}$ & $\mathbf{0.158}$ & $\mathbf{89.865}$ & $\mathbf{9.834}$ & $\mathbf{2.396}$ \\
\midrule
Crab (GT) & $0.367$ & $0.570$ & $0.698$ & $0.019$ & $2.373$ & $12.260$ \\
\midrule
\textbf{Crab (Ours)} & $\mathbf{0.066}$ & $\mathbf{0.109}$ & $\mathbf{0.175}$ & $\mathbf{79.233}$ & $\mathbf{9.862}$ & $\mathbf{2.737}$ \\
\midrule
Cricket (GT) & $0.429$ & $0.658$ & $0.777$ & $0.003$ & $2.015$ & $14.790$ \\
\midrule
\textbf{Cricket (Ours)} & $\mathbf{0.036}$ & $\mathbf{0.075}$ & $\mathbf{0.111}$ & $\mathbf{111.337}$ & $\mathbf{12.126}$ & $\mathbf{3.810}$ \\
\midrule
Crocodile (GT) & $0.567$ & $0.879$ & $0.891$ & $0.016$ & $2.177$ & $13.694$ \\
\midrule
\textbf{Crocodile (Ours)} & $\mathbf{0.054}$ & $\mathbf{0.094}$ & $\mathbf{0.133}$ & $\mathbf{86.886}$ & $\mathbf{10.049}$ & $\mathbf{2.827}$ \\
\midrule
Crow (GT) & $0.390$ & $0.644$ & $0.796$ & $0.018$ & $2.682$ & $12.131$ \\
\midrule
\textbf{Crow (Ours)} & $\mathbf{0.034}$ & $\mathbf{0.058}$ & $\mathbf{0.079}$ & $\mathbf{61.148}$ & $\mathbf{10.915}$ & $\mathbf{5.021}$ \\
\midrule
Deer (GT) & $0.497$ & $0.722$ & $0.816$ & $0.010$ & $2.700$ & $11.116$ \\
\midrule
\textbf{Deer (Ours)} & $\mathbf{0.092}$ & $\mathbf{0.174}$ & $\mathbf{0.231}$ & $\mathbf{12.931}$ & $\mathbf{8.106}$ & $\mathbf{8.577}$ \\
\midrule

Dog (GT) & $0.609$ & $0.850$ & $0.940$ & $0.025$ & $2.001$ & $15.927$ \\
\midrule
\textbf{Dog (Ours)} & $\mathbf{0.039}$ & $\mathbf{0.086}$ & $\mathbf{0.111}$ & $\mathbf{80.711}$ & $\mathbf{12.049}$ & $\mathbf{4.249}$ \\
\midrule
Eagle (GT) & $0.551$ & $0.788$ & $0.896$ & $0.003$ & $1.607$ & $14.266$ \\
\midrule
\textbf{Eagle (Ours)} & $\mathbf{0.057}$ & $\mathbf{0.121}$ & $\mathbf{0.169}$ & $\mathbf{65.728}$ & $\mathbf{10.753}$ & $\mathbf{4.716}$ \\
\midrule
Elephant (GT) & $0.544$ & $0.750$ & $0.851$ & $0.006$ & $2.331$ & $14.017$ \\
\midrule
\textbf{Elephant (Ours)} & $\mathbf{0.044}$ & $\mathbf{0.091}$ & $\mathbf{0.136}$ & $\mathbf{91.635}$ & $\mathbf{10.651}$ & $\mathbf{2.837}$ \\
\midrule
Fire Ant (GT) & $0.399$ & $0.587$ & $0.692$ & $0.004$ & $3.602$ & $9.441$ \\
\midrule
\textbf{Fire Ant (Ours)} & $\mathbf{0.027}$ & $\mathbf{0.055}$ & $\mathbf{0.081}$ & $\mathbf{48.101}$ & $\mathbf{7.988}$ & $\mathbf{2.065}$ \\
\midrule
Flamingo (GT) & $0.078$ & $0.250$ & $0.375$ & $0.023$ & $2.971$ & $8.589$ \\
\midrule
\textbf{Flamingo (Ours)} & $\mathbf{0.031}$ & $\mathbf{0.062}$ & $\mathbf{0.109}$ & $\mathbf{47.836}$ & $\mathbf{8.396}$ & $\mathbf{3.119}$ \\
\midrule
Fox (GT) & $0.380$ & $0.566$ & $0.723$ & $0.009$ & $2.807$ & $11.209$ \\
\midrule
\textbf{Fox (Ours)} & $\mathbf{0.061}$ & $\mathbf{0.128}$ & $\mathbf{0.179}$ & $\mathbf{30.368}$ & $\mathbf{9.234}$ & $\mathbf{7.117}$ \\
\bottomrule
\end{tabular}}
\label{tab:overall}
\end{table}

\begin{table}[htbp]
\centering
\caption{Continued from Table \ref{tab:overall}.}
\vspace{0.5cm}
\resizebox{0.8\columnwidth}{!}{%
\begin{tabular}{lcccccc}
\toprule
Method & R-Prec Top 1 $\uparrow$ & R-Prec Top 2 $\uparrow$ & R-Prec Top 3 $\uparrow$ & FID $\downarrow$ & MultiModal-Dist $\downarrow$ & Diversity $\rightarrow$\\ 
\midrule
Gazelle (GT) & $0.462$ & $0.700$ & $0.825$ & $0.024$ & $2.401$ & $12.907$ \\
\midrule
\textbf{Gazelle (Ours)} & $\mathbf{0.019}$ & $\mathbf{0.041}$ & $\mathbf{0.075}$ & $\mathbf{95.267}$ & $\mathbf{11.966}$ & $\mathbf{4.100}$ \\
\midrule
Giant Bee (GT) & $0.353$ & $0.555$ & $0.703$ & $0.029$ & $2.700$ & $12.992$ \\
\midrule
\textbf{Giant Bee (Ours)} & $\mathbf{0.097}$ & $\mathbf{0.167}$ & $\mathbf{0.234}$ & $\mathbf{63.874}$ & $\mathbf{9.436}$ & $\mathbf{5.031}$ \\
\midrule
Goat (GT) & $0.394$ & $0.663$ & $0.825$ & $0.013$ & $1.975$ & $11.550$ \\
\midrule
\textbf{Goat (Ours)} & $\mathbf{0.033}$ & $\mathbf{0.064}$ & $\mathbf{0.094}$ & $\mathbf{102.515}$ & $\mathbf{10.767}$ & $\mathbf{2.069}$ \\
\midrule
Hamster (GT) & $0.297$ & $0.531$ & $0.672$ & $0.039$ & $2.251$ & $12.088$ \\
\midrule
\textbf{Hamster (Ours)} & $\mathbf{0.052}$ & $\mathbf{0.104}$ & $\mathbf{0.141}$ & $\mathbf{108.243}$ & $\mathbf{10.693}$ & $\mathbf{2.000}$ \\
\midrule
Hermit Crab (GT) & $0.439$ & $0.680$ & $0.811$ & $0.009$ & $2.677$ & $10.232$ \\
\midrule
\textbf{Hermit Crab (Ours)} & $\mathbf{0.046}$ & $\mathbf{0.085}$ & $\mathbf{0.113}$ & $\mathbf{45.140}$ & $\mathbf{8.103}$ & $\mathbf{2.704}$ \\
\midrule
Hippopotamus (GT) & $0.542$ & $0.773$ & $0.879$ & $0.044$ & $2.201$ & $11.970$ \\
\midrule
\textbf{Hippopotamus (Ours)} & $\mathbf{0.052}$ & $\mathbf{0.102}$ & $\mathbf{0.139}$ & $\mathbf{86.651}$ & $\mathbf{9.931}$ & $\mathbf{1.684}$ \\
\midrule
Horse (GT) & $0.518$ & $0.686$ & $0.779$ & $0.006$ & $2.816$ & $10.935$ \\
\midrule
\textbf{Horse (Ours)} & $\mathbf{0.053}$ & $\mathbf{0.090}$ & $\mathbf{0.137}$ & $\mathbf{31.290}$ & $\mathbf{9.131}$ & $\mathbf{6.203}$ \\
\midrule
Hound (GT) & $0.639$ & $0.822$ & $0.909$ & $0.014$ & $2.519$ & $11.514$ \\
\midrule
\textbf{Hound (Ours)} & $\mathbf{0.047}$ & $\mathbf{0.107}$ & $\mathbf{0.146}$ & $\mathbf{68.021}$ & $\mathbf{9.198}$ & $\mathbf{2.529}$ \\
\midrule
Isopetra (GT) & $0.498$ & $0.708$ & $0.824$ & $0.020$ & $2.710$ & $11.242$ \\
\midrule
\textbf{Isoptera (Ours)} & $\mathbf{0.035}$ & $\mathbf{0.088}$ & $\mathbf{0.127}$ & $\mathbf{56.617}$ & $\mathbf{9.598}$ & $\mathbf{3.903}$ \\
\midrule
Jaguar (GT) & $0.538$ & $0.762$ & $0.871$ & $0.009$ & $2.354$ & $11.287$ \\
\midrule
\textbf{Jaguar (Ours)} & $\mathbf{0.043}$ & $\mathbf{0.074}$ & $\mathbf{0.105}$ & $\mathbf{71.047}$ & $\mathbf{9.314}$ & $\mathbf{2.442}$ \\
\midrule
Leopard (GT) & $0.558$ & $0.758$ & $0.884$ & $0.017$ & $1.989$ & $13.777$ \\
\midrule
\textbf{Leopard (Ours)} & $\mathbf{0.058}$ & $\mathbf{0.115}$ & $\mathbf{0.152}$ & $\mathbf{67.369}$ & $\mathbf{9.686}$ & $\mathbf{4.035}$ \\
\midrule
Lion (GT) & $0.542$ & $0.756$ & $0.881$ & $0.010$ & $2.357$ & $11.772$ \\
\midrule
\textbf{Lion (Ours)} & $\mathbf{0.042}$ & $\mathbf{0.066}$ & $\mathbf{0.107}$ & $\mathbf{79.524}$ & $\mathbf{9.760}$ & $\mathbf{2.419}$ \\
\midrule
Lynx (GT) & $0.526$ & $0.751$ & $0.869$ & $0.019$ & $2.158$ & $12.317$ \\
\midrule
\textbf{Lynx (Ours)} & $\mathbf{0.059}$ & $\mathbf{0.096}$ & $\mathbf{0.140}$ & $\mathbf{77.358}$ & $\mathbf{9.666}$ & $\mathbf{2.858}$ \\
\midrule
Mammoth (GT) & $0.551$ & $0.775$ & $0.860$ & $0.004$ & $2.518$ & $12.963$ \\
\midrule
\textbf{Mammoth (Ours)} & $\mathbf{0.055}$ & $\mathbf{0.112}$ & $\mathbf{0.155}$ & $\mathbf{80.510}$ & $\mathbf{9.973}$ & $\mathbf{2.614}$ \\
\midrule
Monkey (GT) & $0.361$ & $0.551$ & $0.683$ & $0.015$ & $3.425$ & $11.441$ \\
\midrule
\textbf{Monkey (Ours)} & $\mathbf{0.053}$ & $\mathbf{0.082}$ & $\mathbf{0.118}$ & $\mathbf{31.473}$ & $\mathbf{8.819}$ & $\mathbf{2.880}$ \\
\midrule
Ostrich (GT) & $0.508$ & $0.691$ & $0.817$ & $0.011$ & $2.507$ & $12.145$ \\
\midrule
\textbf{Ostrich (Ours)} & $\mathbf{0.084}$ & $\mathbf{0.145}$ & $\mathbf{0.189}$ & $\mathbf{59.510}$ & $\mathbf{9.215}$ & $\mathbf{3.988}$ \\
\midrule
Parrot (GT) & $0.524$ & $0.754$ & $0.859$ & $0.040$ & $2.091$ & $14.021$ \\
\midrule
\textbf{Parrot (Ours)} & $\mathbf{0.068}$ & $\mathbf{0.149}$ & $\mathbf{0.202}$ & $\mathbf{27.881}$ & $\mathbf{10.990}$ & $\mathbf{9.306}$ \\
\midrule
Pigeon (GT) & $0.431$ & $0.670$ & $0.810$ & $0.047$ & $2.151$ & $11.879$ \\
\midrule
\textbf{Pigeon (Ours)} & $\mathbf{0.208}$ & $\mathbf{0.328}$ & $\mathbf{0.426}$ & $\mathbf{0.431}$ & $\mathbf{7.187}$ & $\mathbf{11.446}$ \\
\midrule
Piranha (GT) & $0.287$ & $0.499$ & $0.684$ & $0.010$ & $2.262$ & $11.645$ \\
\midrule
\textbf{Piranha (Ours)} & $\mathbf{0.038}$ & $\mathbf{0.069}$ & $\mathbf{0.096}$ & $\mathbf{81.586}$ & $\mathbf{9.930}$ & $\mathbf{1.719}$ \\
\midrule
Polar Bear (GT) & $0.519$ & $0.746$ & $0.867$ & $0.024$ & $2.162$ & $12.509$ \\
\midrule
\textbf{Polar Bear (Ours)} & $\mathbf{0.031}$ & $\mathbf{0.065}$ & $\mathbf{0.102}$ & $\mathbf{112.532}$ & $\mathbf{11.160}$ & $\mathbf{3.155}$ \\
\midrule
Pteranodon (GT) & $0.510$ & $0.730$ & $0.815$ & $0.021$ & $2.704$ & $10.532$ \\
\midrule
\textbf{Pteranodon (Ours)} & $\mathbf{0.037}$ & $\mathbf{0.076}$ & $\mathbf{0.121}$ & $\mathbf{26.197}$ & $\mathbf{8.329}$ & $\mathbf{1.424}$ \\
\midrule
Puppy (GT) & $0.344$ & $0.563$ & $0.729$ & $0.036$ & $1.421$ & $12.181$ \\
\midrule
\textbf{Puppy (Ours)} & $\mathbf{0.094}$ & $\mathbf{0.240}$ & $\mathbf{0.292}$ & $\mathbf{55.135}$ & $\mathbf{8.730}$ & $\mathbf{5.339}$ \\
\midrule

Raptor (GT) & $0.531$ & $0.781$ & $0.881$ & $0.066$ & $1.896$ & $13.216$ \\
\midrule
\textbf{Raptor (Ours)} & $\mathbf{0.077}$ & $\mathbf{0.151}$ & $\mathbf{0.216}$ & $\mathbf{105.263}$ & $\mathbf{10.961}$ & $\mathbf{4.023}$ \\
\midrule
Rat (GT) & $0.396$ & $0.639$ & $0.802$ & $0.057$ & $2.609$ & $11.625$ \\
\midrule
\textbf{Rat (Ours)} & $\mathbf{0.243}$ & $\mathbf{0.385}$ & $\mathbf{0.482}$ & $\mathbf{0.495}$ & $\mathbf{6.247}$ & $\mathbf{11.592}$ \\
\midrule
Reindeer (GT) & $0.580$ & $0.833$ & $0.918$ & $0.020$ & $2.177$ & $14.077$ \\
\midrule
\textbf{Reindeer (Ours)} & $\mathbf{0.043}$ & $\mathbf{0.073}$ & $\mathbf{0.106}$ & $\mathbf{130.645}$ & $\mathbf{11.970}$ & $\mathbf{2.953}$ \\
\bottomrule
\end{tabular}}
\label{tab:overall2}
\end{table}

\begin{table}[htbp]
\centering
\caption{Continued from Table \ref{tab:overall2}.}
\vspace{0.5cm}
\resizebox{0.85\columnwidth}{!}{%
\begin{tabular}{lcccccc}
\toprule
Method & R-Prec Top 1 $\uparrow$ & R-Prec Top 2 $\uparrow$ & R-Prec Top 3 $\uparrow$ & FID $\downarrow$ & MultiModal-Dist $\downarrow$ & Diversity $\rightarrow$\\
\midrule
Rhino (GT) & $0.349$ & $0.575$ & $0.690$ & $0.031$ & $2.971$ & $10.070$ \\
\midrule
\textbf{Rhino (Ours)} & $\mathbf{0.055}$ & $\mathbf{0.106}$ & $\mathbf{0.147}$ & $\mathbf{60.834}$ & $\mathbf{8.572}$ & $\mathbf{3.423}$ \\
\midrule
Roach (GT) & $0.516$ & $0.693$ & $0.828$ & $0.014$ & $2.456$ & $12.001$ \\
\midrule
\textbf{Roach (Ours)} & $\mathbf{0.047}$ & $\mathbf{0.120}$ & $\mathbf{0.182}$ & $\mathbf{56.589}$ & $\mathbf{9.980}$ & $\mathbf{5.248}$ \\
\midrule
Sabre-toothed tiger (GT) & $0.625$ & $0.795$ & $0.867$ & $0.029$ & $2.804$ & $10.963$ \\
\midrule
\textbf{Sabre-toothed Tiger (Ours)} & $\mathbf{0.035}$ & $\mathbf{0.071}$ & $\mathbf{0.098}$ & $\mathbf{18.381}$ & $\mathbf{8.898}$ & $\mathbf{5.478}$ \\
\midrule
Sand mouse (GT) & $0.373$ & $0.595$ & $0.742$ & $0.007$ & $2.878$ & $11.316$ \\
\midrule
\textbf{Sand Mouse (Ours)} & $\mathbf{0.035}$ & $\mathbf{0.067}$ & $\mathbf{0.118}$ & $\mathbf{65.683}$ & $\mathbf{9.128}$ & $\mathbf{3.479}$ \\
\midrule
Scorpion (GT) & $0.548$ & $0.772$ & $0.888$ & $0.022$ & $1.702$ & $12.928$ \\
\midrule
\textbf{Scorpion (Ours)} & $\mathbf{0.048}$ & $\mathbf{0.809}$ & $\mathbf{0.108}$ & $\mathbf{63.289}$ & $\mathbf{10.349}$ & $\mathbf{4.709}$ \\
\midrule
Shark (GT) & $0.412$ & $0.627$ & $0.760$ & $0.022$ & $3.118$ & $9.744$ \\
\midrule
\textbf{Shark (Ours)} & $\mathbf{0.210}$ & $\mathbf{0.335}$ & $\mathbf{0.387}$ & $\mathbf{0.568}$ & $\mathbf{7.090}$ & $\mathbf{9.415}$ \\
\midrule
Skunk (GT) & $0.484$ & $0.712$ & $0.786$ & $0.021$ & $2.615$ & $11.999$ \\
\midrule
\textbf{Skunk (Ours)} & $\mathbf{0.022}$ & $\mathbf{0.056}$ & $\mathbf{0.094}$ & $\mathbf{81.971}$ & $\mathbf{10.250}$ & $\mathbf{3.155}$ \\
\midrule
Spider (GT) & $0.373$ & $0.618$ & $0.752$ & $0.007$ & $2.279$ & $12.354$ \\
\midrule
\textbf{Spider (Ours)} & $\mathbf{0.060}$ & $\mathbf{0.103}$ & $\mathbf{0.150}$ & $\mathbf{81.980}$ & $\mathbf{10.375}$ & $\mathbf{2.722}$ \\
\midrule
Stegosaurus (GT) & $0.371$ & $0.573$ & $0.706$ & $0.049$ & $2.976$ & $10.145$ \\
\midrule
\textbf{Stegosaurus (Ours)} & $\mathbf{0.042}$ & $\mathbf{0.073}$ & $\mathbf{0.113}$ & $\mathbf{59.360}$ & $\mathbf{7.915}$ & $\mathbf{1.001}$ \\
\midrule
T-Rex (GT) & $0.125$ & $0.250$ & $0.344$ & $0.163$ & $4.367$ & $5.915$ \\
\midrule
\textbf{T-Rex (Ours)} & $\mathbf{0.000}$ & $\mathbf{0.031}$ & $\mathbf{0.063}$ & $\mathbf{31.338}$ & $\mathbf{7.166}$ & $\mathbf{2.823}$ \\
\midrule
Tricera (GT) & $0.316$ & $0.502$ & $0.625$ & $0.004$ & $3.626$ & $10.532$ \\
\midrule
\textbf{Tricera (Ours)} & $\mathbf{0.047}$ & $\mathbf{0.090}$ & $\mathbf{0.137}$ & $\mathbf{62.515}$ & $\mathbf{8.280}$ & $\mathbf{1.059}$ \\
\midrule
Toucan (GT) & $0.489$ & $0.723$ & $0.842$ & $0.012$ & $2.457$ & $13.668$ \\
\midrule
\textbf{Toucan (Ours)} & $\mathbf{0.205}$ & $\mathbf{0.364}$ & $\mathbf{0.464}$ & $\mathbf{0.915}$ & $\mathbf{7.441}$ & $\mathbf{12.437}$ \\
\midrule
Turtle (GT) & $0.426$ & $0.665$ & $0.790$ & $0.016$ & $2.266$ & $11.510$ \\
\midrule
\textbf{Turtle (Ours)} & $\mathbf{0.036}$ & $\mathbf{0.078}$ & $\mathbf{0.123}$ & $\mathbf{64.208}$ & $\mathbf{8.913}$ & $\mathbf{2.299}$ \\
\midrule
Tyrannosaurus Rex (GT) & $0.425$ & $0.681$ & $0.799$ & $0.007$ & $1.606$ & $11.587$ \\
\midrule
\textbf{Tyrannosaurus Rex (Ours)} & $\mathbf{0.054}$ & $\mathbf{0.087}$ & $\mathbf{0.144}$ & $\mathbf{68.680}$ & $\mathbf{8.512}$ & $\mathbf{1.638}$ \\
\midrule
Wyvern (GT) & $0.424$ & $0.643$ & $0.750$ & $0.077$ & $2.693$ & $11.574$ \\
\midrule
\textbf{Wyvern (Ours)} & $\mathbf{0.098}$ & $\mathbf{0.196}$ & $\mathbf{0.290}$ & $\mathbf{43.339}$ & $\mathbf{8.184}$ & $\mathbf{4.517}$ \\
\midrule
\midrule
Average (GT) & $0.437$ & $0.650$ & $0.767$ & $0.024$ & $2.559$ & $11.790$ \\
\midrule
\textbf{Average (Ours)} & $\mathbf{0.069}$ & $\mathbf{0.144}$ & $\mathbf{0.173}$ & $\mathbf{62.785}$ & $\mathbf{9.371}$ & $\mathbf{4.158}$ \\
\bottomrule
\end{tabular}}
\label{tab:overall3}
\end{table}

\section{User Study}
\label{app:b}
\begin{figure}[ht]
    \centering
    \includegraphics[width=0.7\linewidth]{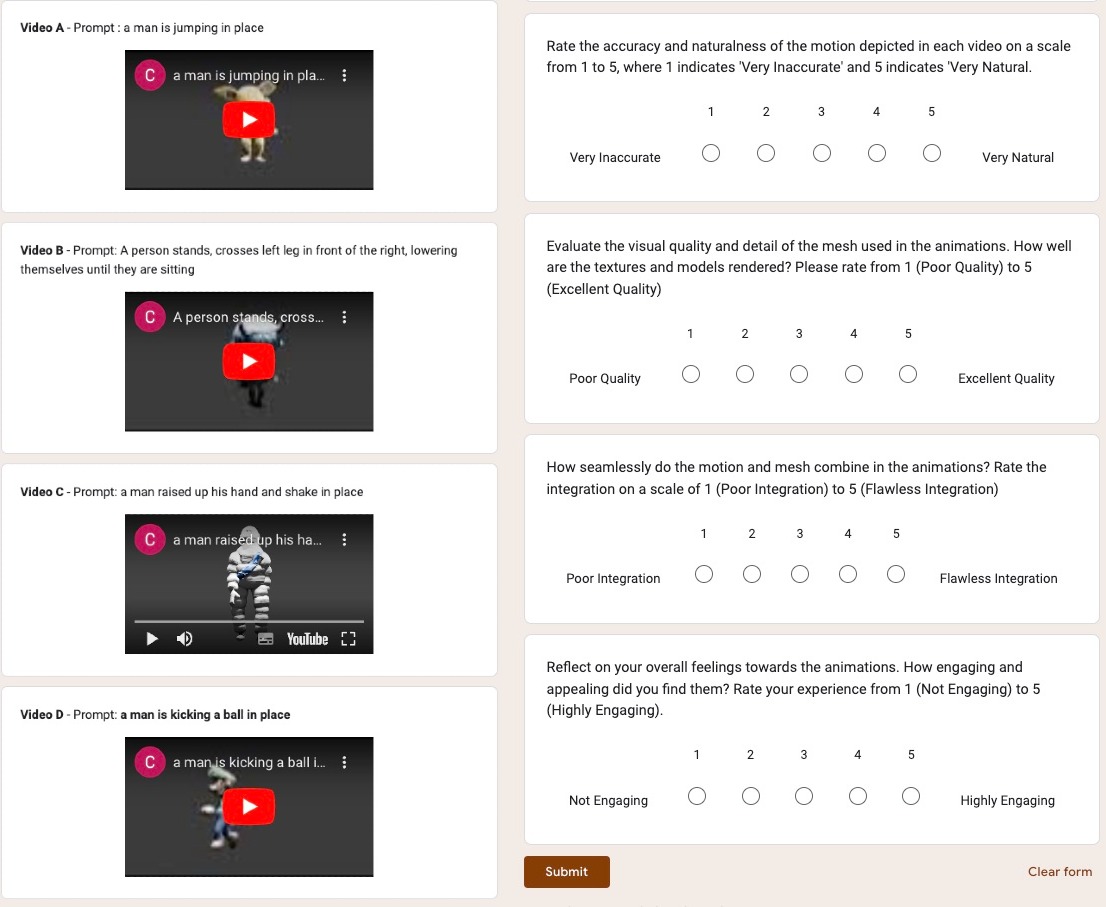}
    \vspace{20pt} 
    \caption{This figure displays the User Interface (UI) used in our User Study, showcasing four videos (Video A to D) each with distinct motion animations from various models. Participants evaluate these animations on aspects such as motion accuracy, mesh quality, integration of motion and mesh, and overall user experience. They rate each aspect from 1 (low) to 5 (high) to assess how the animations mirror real-world movements, the visual appeal of the animations, their integration quality, and their engagement level. This evaluation aims to determine the realism and engagement effectiveness of each animation model.}
    \label{fig:user_study}
\end{figure}

In this work, we conduct an exhaustive evaluation of the effectiveness of our motion avatar generation, leveraging both qualitative and quantitative assessments. This user study assesses the real-world applicability of 4 motion video generated from the Motion Avatar platform by the input prompt, examined by 50 participants through a Google Forms interface as in Fig~\ref{fig:user_study}.

Participants were presented with four videos labeled Video A, Video B, Video C, and Video D. Each video showcased a unique motion generated using Motion Avatar with different input prompt. The participants evaluated these videos by responding to a series of targeted questions aimed at assessing the motion's accuracy, the mesh's visual quality, the integration of motion and mesh, and their overall emotional response to the animations.

The evaluation was structured around several key aspects:

\begin{enumerate}

    \item Motion Accuracy: Participants rated the naturalness and accuracy of the motions on a scale from 1 ('Very Inaccurate') to 5 ('Very Natural'). The average score was 4.2, indicating a high fidelity in motion portrayal.
    \item Mesh Quality: The visual quality and detail of the mesh were rated from 1 ('Poor Quality') to 5 ('Excellent Quality'), with an average score of 4.0, highlighting the superior visual appeal of our models.
    \item Motion and Mesh Integration: The integration of motion and mesh was assessed, with most participants rating this aspect a 4.5 on average, reflecting seamless integration that enhances fluidity and realism.
    \item User Engagement and Appeal: Participants reflected on their feelings towards the animations, rating their overall engagement and appeal from 1 ('Not Engaging') to 5 ('Highly Engaging'). The average engagement score was 4.3, suggesting that the animations were highly engaging and appealing to the audience.
\end{enumerate}

Results suggested that: 
\begin{itemize}
    \item 92\% of participants believed the animations could be directly utilized in real-world applications without significant modifications.
    \item Only 8\% felt that minor adjustments were necessary before deployment.
\end{itemize}

These findings underscore the high quality, impressive results, and broad usability of the animations. Nearly all participants found the quality of the generated videos to be very high, with no significant criticisms regarding their quality. The results were notably impressive, with many participants expressing enthusiasm for widespread use of these animations.

This comprehensive user study confirms that our animations not only meet but exceed user expectations in terms of quality, realism, and engagement, making them highly suitable for varied practical applications. The insights from this study will guide further enhancements to ensure our animation generation remains at the forefront of technological and artistic innovation.

\section{Avatar Q\&A Dataset Example}
\label{app:c}

\begin{center}
\begin{minipage}{8cm}
\centering
\begin{lstlisting}[language=json, basicstyle=\ttfamily\bfseries\mycustomsize, linewidth=8cm, caption={Examples from the \textbf{Avatar Q\&A Dataset}}, label=lst:qa]
[
    {
        "instruction": "I saw an animal attacking something. A closer look makes it clear that a monkey is attacking",
        "input": "",
        "output": "The animal is {Monkey}, and motion is {Attack}.",
        "history": []
    },
    {
        "instruction": "A chicken walked quickly from my line of sight",
        "input": "",
        "output": "The animal is {Chicken}, and motion is {Walk Quick}.",
        "history": []
    },
    {
        "instruction": "A fox, seeing the food ahead, lowers its body and slowly approaches its prey",
        "input": "",
        "output": "The animal is {Fox}, and motion is {Low Bite}.",
        "history": []
    },
    {
        "instruction": "A fox walked out of the woods.",
        "input": "",
        "output": "The animal is {Fox}, and motion is {Walk Out}.",
        "history": []
    },
    {
        "instruction": "The rabbit hopped across the meadow, its fluffy tail bouncing in the sunlight.",
        "input": "",
        "output": "The animal is {Rabbit}, and motion is {Hop}.",
        "history": []
    }
]
\end{lstlisting}
\end{minipage}
\end{center}

\end{document}